\documentclass{article}
\usepackage{spconf,amsmath,graphicx,amssymb}

\usepackage{enumitem}
\setlist{nosep, leftmargin=14pt}

\usepackage{mwe} 

\title{Cardiac Motion Modeling with Parallel Transport and Shape Splines}
\name{Nicolas Guigui$^{\dagger}$ \quad 
    Pamela Moceri$^{\star \dagger}$ \quad
    Maxime Sermesant$^{\dagger}$ \quad
    Xavier Pennec$^{\dagger}$ \quad}

\address{$^{\dagger}$Universit\'e C\^ote d'Azur, Inria, Epione, Sophia-Antipolis, France \\
    $^{\star}$UR2CA, Université C\^ote d’Azur, Facult\'e de M\'edecine, Nice, France}

\begin{document}
%
\maketitle
\begin{abstract}
In cases of pressure or volume overload, probing cardiac function may be difficult because of the interactions between shape and deformations.
In this work, we use the LDDMM framework and parallel transport to estimate and reorient deformations of the right ventricle. 
We then propose a normalization procedure for the amplitude of the deformation, and a second-order spline model to represent the full cardiac contraction.
The method is applied to 3D meshes of the right ventricle extracted from echocardiographic sequences of 314 patients divided into three disease categories and a control group. 
We find significant differences between pathologies in the model parameters, revealing insights into the dynamics of each disease. 
\end{abstract}
\section{Introduction}
\label{sec:intro}

Spatio-temporal shape analysis is of growing importance in the study of cardiac diseases.
In particular, the assessment of cardiac function requires the measurement and analysis of cardiac motion beyond scalar indicators such as ejection fraction or area strain. The case of the Right Ventricle (RV), is of particular interest as it has been shown to have a large capacity to adapt to overload by remodelling~\cite{sanz_anatomy_2019}, raising the issue of disentangling the deformation from the initial anatomy.

A powerful and popular viewpoint based on shape registration is that of Riemannian Geometry and the Large Deformations Diffeomorphic Metric Mapping (LDDMM) framework, modelling non-linear transformations as elements of the group of diffeomorphisms of the embedding space, to ensure smooth and invertible mappings between shapes, and using parallel transport to build a reference-centred representation of all the patients' trajectories. Parallel transport is a natural geometric tool to normalise deformations that is consistent with registration.
Two main approaches have been proposed. The first consists in estimating a mean trajectory, and representing each individual's trajectory as a translation of this mean evolution by a subject-specific perturbation, that is parallel transported to each time point \cite[and references therein]{schiratti_bayesian_2017}. The other strategy \cite[and references therein]{lorenzi_efficient_2014,younes_shapes_2019}, which will be developed in this paper, consists in first representing each trajectory with respect to its own reference, and then transporting it to a global reference --the atlas-- thus normalizing the individual deformations.

However, the computation of parallel transport in the diffeomorhism group is challenging, as no closed form solutions exist and the numerical methods lack stability or result in over-smoothing.
Moreover, \cite{niethammer_riemannian_2013} showed that parallel transport in LDDMM does not conserve global properties such as scale or volume changes. In the case of cardiac deformations, the magnitude of the temporal deformation is comparable to that of the subject's reference to atlas deformation, and substantial volume changes are observed. Thus the lack of scale-invariance is crucial. 
In this work, we investigate the effect of a straightforward scaling of the transported deformation, in order to preserve the ejection fraction of the RV.

Furthermore, we leverage recent results on the Pole Ladder \cite{guigui_numerical_2020}, a method that consists in approximating the parallel transport of a tangent vector along a geodesic. 
This method was first proposed by \cite{lorenzi_efficient_2014}, for brain data in the context of Alzheimer's disease.
We use it, in the context of LDDMM, to associate a deformation of the atlas to each shape. This allows to build patient-specific trajectories all starting from the atlas.

The patient-specific trajectories are then summarized by fitting a spline regression, which postulates a second-order dynamic model \cite{trouve_shape_2012} 
whose parameters are estimated with an optimization procedure. This allows to compactly represent all the patients' trajectories in the same space, and to proceed with linear statistics.

\vspace{-2mm}
\section{Motivation: the Right Ventricle under pressure}
\label{sec:motivation}

With this framework, we study the RV under pressure or volume overload due to different diseases: Pulmonary Hypertension (PHT), Tetralogy of Fallot (ToF) and Atrial Septal Defect (ASD) and seek to characterize their impact on the contraction of the RV, during the systolic phase of the cardiac cycle. Previous work on these pathologies demonstrated differences in RV function between the ASD group and the ToF group, despite comparable shape remodelling~\cite{moceri_3d_2020}. Moreover, \cite{moceri_three-dimensional_2018} showed that the Area Strain (AS), i.e. the relative change of area of each cell of the mesh that represents the RV, was a strong predictor of survival in the PHT group. \cite{di_folco_learning_2019} studied the interactions between AS and shape descriptors.

Both studies suffered from a low power due to a small cohort against high-dimensional markers. Thanks to the use of sparse control points to parameterize the LDDMM and spline deformations, our descriptors are more compact and translate into increased statistical power.

We use 3D meshes extracted from 314 echocardiographic sequences from patients examined at the CHU of Nice. The meshes were extracted with a commercial software (4D RV Function 2.0, TomTec Imaging Systems, GmbH, DE) with point-to-point correspondences across time and patients. These are formed by 938 points and 1872 triangles. All the shapes were realigned with a subject-specific rigid-body deformation. An atlas was computed from the end diastolic meshes of the control group, after alignment.

\section{Method}
\label{sec:method}
\subsection{The LDDMM framework}
\label{sec:lddmm}
The LDDMM framework encompasses both algorithms for shape matching (a.k.a.\ registration) and a Riemannian geometric structure on the space of shapes. The former allows to compute shape descriptors, and to parameterize diffeomorphisms. The latter, provides a distance to compare deformations, and an associated notion of parallelism to transport them.
We present here the formulation of \cite{durrleman_morphometry_2014} and its implementation in \cite{bone_deformetrica_2018}, focusing on the case of landmarks. 

A natural and efficient computational construction of diffeomorphisms is obtained by flows associated to ordinary differential equations (ODEs) $\partial \phi_t(\cdot) = v_t[\phi_t(\cdot)]$, with the initial condition $\phi_0 = Id$. The time-dependent vector field $v_t$ can be interpreted as the instantaneous speed of the points during deformation, and must verify certain regularity conditions to ensure that solutions to the ODE are indeed diffeomorphisms. An efficient way to enforce these conditions is to consider vector fields obtained by the convolution of a number $N_c$ of momentum vectors carried by control points: $v_t(x) = \sum_{k=1}^{N_c} K(x, c^{(t)}_k) \mu^{(t)}_k$, where $K$ is the Gaussian kernel: $K(x,y) = \exp (- \frac{\Vert x-y \Vert^2}{\sigma^2})$. The (closure of the) set of such vector fields forms a reproducing kernel Hilbert space, with the associated norm $\|v\|_K^2 = \sum_{i,j} K(c_i, c_j)\mu_i^T \mu_j$. The total cost, or energy of the deformation can be defined as $\int_0^1 \|v_t\|_K^2 dt$.

It can be shown that the momentum vectors that minimize this energy, considering $c^{(0)}_k,c^{(1)}_k, \; k=1\ldots N_c$ fixed, together with the equation driving the motion of the control points, follow a Hamiltonian system of ODEs:
\begin{equation}
    \begin{cases}
    \Dot{c_k}^{(t)} &= \sum_j K(c_k^{(t)}, c_j^{(t)}) \mu_j^{(t)} \\
    \Dot{\mu_k}^{(t)} &= - \sum_j \nabla_1 K(c_k^{(t)}, c_j^{(t)}) \mu_k^{(t)^T}\mu_j^{(t)}
    \end{cases}
    \label{eq:hamilt_sys}
\end{equation}
A diffeomorphism $\phi_1$ is thus uniquely parameterized by the initial conditions $c_k^{(0)}, \mu_k^{(0)}, \; k=1\ldots N_c$, and a shape registration criterion between a template $T$ and target $S$ can be defined as
\begin{equation}
    C(c, \mu) = \Vert S - \phi_1^{c,\mu}(T) \Vert_2^2 + \alpha^2 \| v_0^{c,\mu} \|_K^2.
    \label{registration_crit}
\end{equation}
where $\alpha$ is a regularisation parameter that penalises large deformations. Minimizing $C$ therefore amounts to finding the transformation that best deforms $T$ to match $S$. The gradient of $C$ can be computed through automatic differentiation, to perform gradient descent. The optimal value of $C$ defines a distance between $\phi_1$ and the identity. In fact this distance derives from an invariant Riemannian metric on the group of diffeomorphisms and the path $\phi_t$ is a minimizing geodesic for this metric. By considering the action of diffeomorphisms on shapes, it projects to a distance between the shapes $S$ and $T$.

\subsection{Scaled parallel transport with the pole ladder}
Along with a distance, the Riemannian metric provides a notion of parallel transport. It is defined by an ODE that allows to transport a set of momentum vectors along a path of diffeomorphisms $\phi_t$ (\cite[section 13.3.3]{younes_shapes_2019}).
However, this differential equation is hard to solve in practice and alternative methods have been proposed for the case of transporting along a geodesic.
We here leveraged recent results on the convergence properties of the Pole Ladder \cite{guigui_numerical_2020} to propose a new implementation within the LDDMM framework. 
We solve the registration problem~\eqref{eq:regression_crit} between the end-diastolic (ED) shape and each time frame $t_i$, then use the Pole Ladder to transport this deformation to the atlas, and reconstruct a corresponding shape at time $t_i$.

\begin{figure}[tb]
    \centering
    \includegraphics[width=8cm]{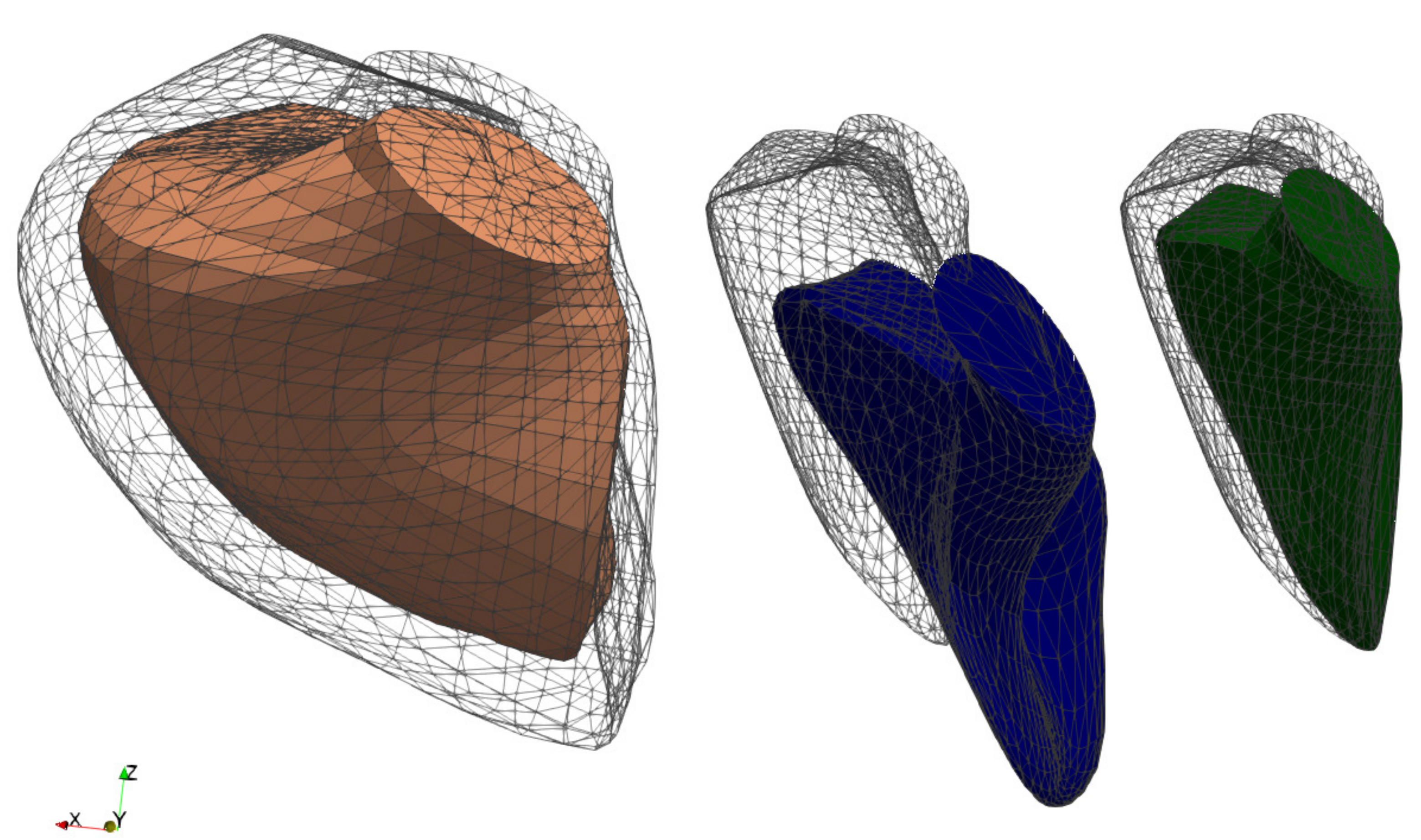}
    \vspace{-2mm}
    \caption{Example: transparent ED mesh over coloured ES mesh. Left: Original patient with extreme volume, Middle: after parallel transport, unrealistic deformation, Right: after scaling of the parallel transport.}
    \label{fig:ex}
    \vspace{-2mm}
\end{figure}

However, there is a substantial correlation between the magnitude of the systolic deformation and the End Diastolic (ED) volume ($\rho=0.42$ in the data-set considered in this paper). As the parallel transport is isometric, this deformation may be too large for the atlas. An example of the obtained end-systolic (ES) frame is shown on Figure~\ref{fig:ex} for a patient whose RV volume is greater that that of the atlas, which results in an unrealistic ES frame. A clinically relevant quantity that should be conserved is the Ejection Fraction (EF), defined as the relative volume change, as it is a straightforward indicator of cardiac function. We then introduce a parameter $\lambda$ such that scaling the magnitude of each intra-subject deformation after parallel transport conserves the EF relative to the ED frame. This parameter is optimized by gradient descent on each patient.
We validate this scaling by probing the conservation of the area strain, and the ES ejection fraction.

\subsection{Spline regression for shape evolution}
The registration framework described in section~\ref{sec:lddmm} estimates a geodesic path between two shapes (the equivalent of a uniform motion). For a trajectory such as the contraction of the cardiac RV, one may expect to find second-order dynamics, making a (first-order) geodesic regression ill-suited. We thus propose to use the second-order model defined in \cite{trouve_shape_2012} to account for the motion of the RV during systole. The second-order terms $u_t$ can be interpreted as random external forces smoothly perturbing the trajectory around a mean geodesic. They modify the continuous-time system of equations~\eqref{registration_crit} as follows: $\forall t \in [0,1]$,
\begin{equation}
    \begin{cases}
    \Dot{c_k}^{(t)} &= \sum_j K(c_k^{(t)}, c_j^{(t)}) \mu_j^{(t)} \\
    \Dot{\mu_k}^{(t)} &= - \sum_j \nabla_1 K(c_k^{(t)}, c_j^{(t)}) \mu_k^{(t)^T}\mu_j^{(t)} + u_k^{(t)}
    \end{cases}
    \label{eq:spline_sys}
\end{equation}

If we consider a discrete sequence of observation times $t_1=0,t_2, \ldots t_d=1$ and configurations $x_{t_1}, \ldots, x_{t_d}$, one seeks to find the path $\phi_t$ that minimizes the new cost
\begin{align}
\vspace{-4mm}
    C_S(c, \mu, u_t) &=  \frac{1}{\alpha^2d} \sum_{i=1}^d\| x_{t_i} - \phi_{t_i}(x_{t_0}) \|_2^2 \nonumber\\
                    &\qquad + \int_0^1 \|u^{(t)}\|^2 dt + \|v_0^{c,\mu}\|_K^2.
    \label{eq:regression_crit}
\end{align}
In practice, the ODEs~\eqref{eq:hamilt_sys} and \eqref{eq:spline_sys} are discretized in $n$ time steps and an integration method such as Euler or Runge-Kutta is used. We define all the patients trajectories between $t=0$ and $t=1$, and use the same discretization for all the patients to ensure that $u_{0},\ldots, u_{N_c}$ are estimated at corresponding times. Along with $\mu^{(0)}$, these are estimated by gradient descent as in the case of registration. We use a kernel bandwidth $\sigma = 15$ in all the experiments, and $60$ control points for all the deformations of the atlas.
The initial control points are fixed for the entire data-set so that the initial momenta can be compared consistently. They have been optimized to register the atlas on all the transported ES frames.

\section{Results}
\label{sec:results}
\subsection{AS and EF conservation}
\begin{figure}
    \centering
    \includegraphics[width=8cm]{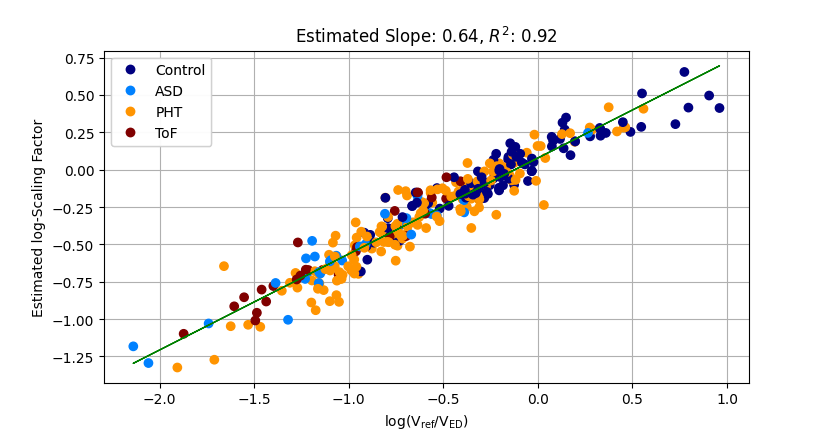}
    \vspace{-4mm}
    \caption{Scaling Parameter $\lambda$ with respect to the ED volume.}
    \label{fig:scaling}
    \vspace{-2mm}
\end{figure}

As expected, the scaling coefficient is closely related to the ED volume. We use a linear regression to identify a linear relation between $\mathrm{log}(\lambda)$ and $\mathrm{log(V_{ref}/V_{ED})}$. This is displayed on Figure~\ref{fig:scaling}.

We validate our scaled parallel transport algorithm by assessing the conservation of scalar quantities of interest: the ejection fraction (EF) and area strain (AS). Both quantities are defined as the relative change of respectively volume or area during systole. The AS is a local quantity computed on each cell of the mesh. We compute the root mean squared error (RMSE) at each cell, and report here the mean over all cells.

We compute the EF on the original data, and compare it with the values computed on the shapes obtained by parallel transport (PT), and scaled parallel transport (SPT). The RMSE are displayed on Table~\ref{tab:ef-as}. Interestingly, it is possible to obtain a low error on the EF after the scaled transport, but this does not preserve the AS. This shows that although these quantities are related, they carry different information and that AS depends on the initial shape, itself related to the pathology. 

\begin{table}[htb]
    \centering
    \begin{tabular}{c|c|c|c}
         &  Original Values & RMSE PT & RMSE SPT \\
         \hline
        AS & -0.24 $\pm$ 0.08 & 0.18 & 0.13 \\
        EF & 0.42 $\pm$ 0.13 & 0.13 & 0.04
    \end{tabular}
    \caption{Validation of parallel transport with EF and AS.}
    \label{tab:ef-as}
    \vspace{-7mm}
\end{table}

\subsection{Groupwise differences on the splines}

\begin{figure}
    \centering
    \includegraphics[width=7cm]{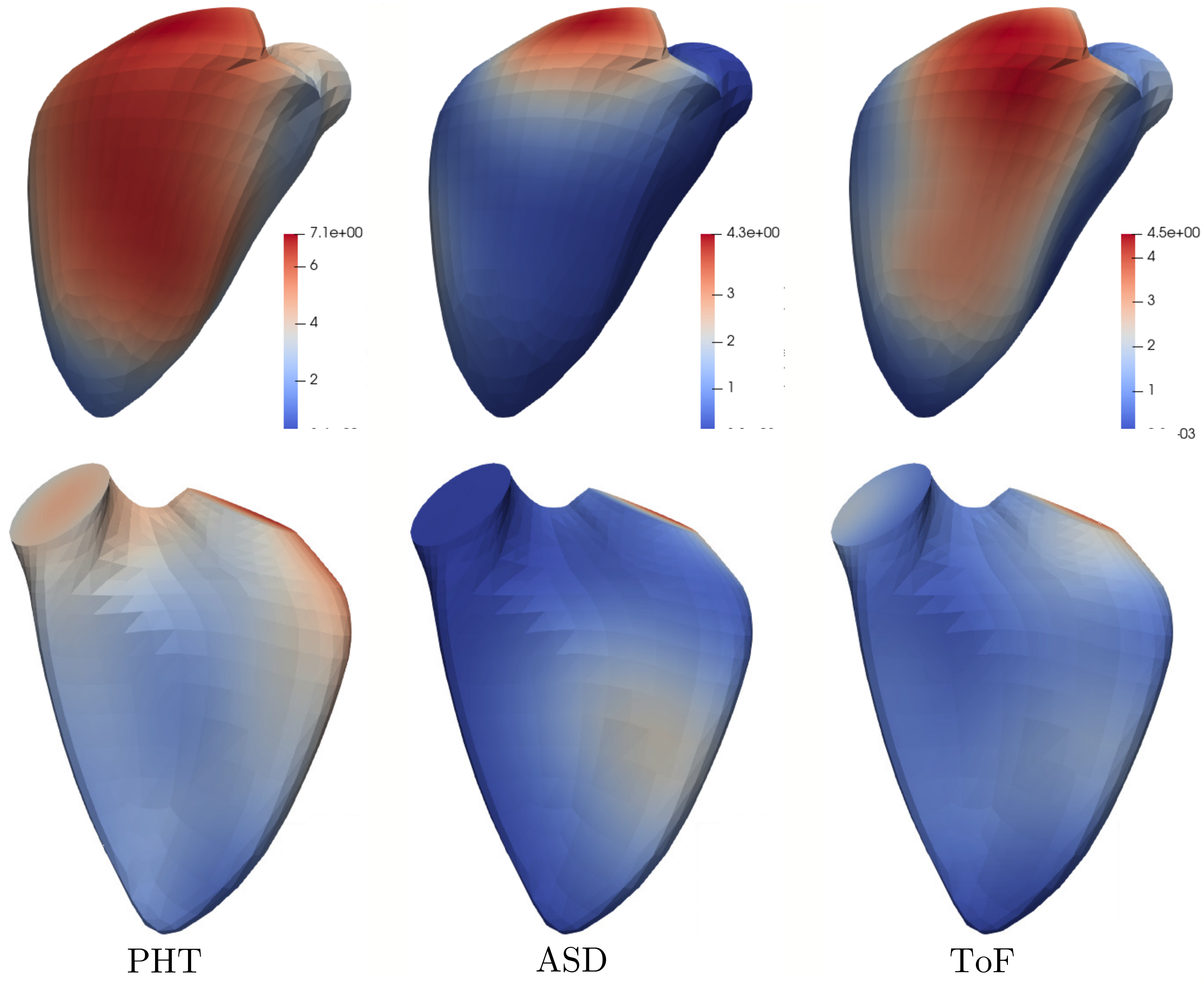}
    \vspace{-3mm}
    \caption{Group Mean momentum where it significantly differs from the control group. Red indicates stronger impact, while blue indicates no impact (no difference). Top row: free wall. Bottom row: septum}
    \label{fig:hotelling}
    \vspace{-3mm}
\end{figure}

Secondly, we study the differences between the diseases on the spline deformations. These are parameterized by the initial momentum $\mu$, and by the discretized external forces $(u^{(0)}, \ldots, u^{(1)})$. We perform a Hotelling multivariate test to compare each disease to the control group, and perform a Bonferroni correction for multiple testing, to maintain type~I error risk at $\alpha=0.05$. The results for the momentum are displayed Figure~\ref{fig:hotelling} for the ASD, ToF and PHT groups and show significant differences between each disease and the control group. The differences observed near the tricuspid valve mainly reflect the difference of magnitude of the deformations, and one should be cautions before drawing further conclusions as the quality of the mesh may vary near this region. However it is interesting to notice that very little differences are observed for the ASD group, which corroborates previous results \cite{moceri_3d_2020}. Similarly, only small differences are observed on the septum, showing that the shape differences usually observed on the PHT group have been filtered out. This makes the differences observed on the free wall interesting and other markers such as the circumferential strain will be studied to confirm these effects.

The second-order terms give insight into the dynamic differences between the groups. Indeed, significant differences were found again for the ToF and PHT groups, but the locations and orientations vary across time. This will be studied more in depth in future work.

\vspace{-3mm}
\section{Compliance with Ethical Standards}
\label{sec:ethics}
This study was performed in line with the principles of
the Declaration of Helsinki. Approval was granted by the CPP Sud M\'editerran\'ee V (2017-A02077-46).

\section{Acknowledgments}
\label{sec:acknowledgments}
This work was partially funded by the ERC grant Nr. 786854 G-Statistics from the European Research Council under the European Union’s Horizon 2020 research and innovation program.
It was also supported by the French government through the 3IA Côte d’Azur Investments ANR-19-P3IA-0002 managed by the National Research Agency. 
The authors are grateful to the OPAL infrastructure from Université Côte d'Azur for providing resources and support. They warmly thank Nicolas Duchateau for his preparatory work on the data and his helpful comments. The authors have no conflict of interest to declare.
\bibliographystyle{IEEEtrans}
\bibliography{ISBI21}

\end{document}